\documentclass[letterpaper, 10 pt, conference]{ieeeconf}  %

\IEEEoverridecommandlockouts                              %

\usepackage{booktabs}       %
\usepackage{amsfonts}       %
\usepackage{amsmath,amssymb}
\usepackage{graphicx}
\usepackage{multirow}
\usepackage[export]{adjustbox}
\usepackage{xcolor}
\usepackage[]{overpic}

\makeatletter
\let\NAT@parse\undefined
\makeatother

\usepackage[breaklinks=true,colorlinks,bookmarks=false]{hyperref}
\usepackage[capitalise]{cleveref}

\newcommand\Tstrut{\rule{0pt}{2.6ex}}       %
\usepackage{url}

\renewcommand{\baselinestretch}{0.94}

\title{\LARGE \bf
Propagating State Uncertainty Through Trajectory Forecasting
}

\author{Boris Ivanovic$^{1\dagger}$, Yifeng (Richard) Lin$^{1\dagger}$, Shubham Shrivastava$^{2}$, Punarjay Chakravarty$^{2}$, Marco Pavone$^{1,3}$%
\thanks{*This work was supported in part by the Ford-Stanford Alliance as well as the Natural Sciences and Engineering Research Council of Canada (NSERC), funding reference number 545934-2020. This article solely reflects the opinions and conclusions of its authors.}%
\thanks{$^{\dagger}$This work was completed while the authors were at Stanford University.}%
\thanks{$^{1}$Boris Ivanovic and Yifeng (Richard) Lin are with NVIDIA
        {\tt\small \{bivanovic, yifengl\}@nvidia.com}}%
\thanks{$^{2}$Shubham Shrivastava and Punarjay Chakravarty are with Ford Greenfield Labs 
        {\tt\small \{sshriva5, pchakra5\}@ford.com}}%
\thanks{$^{3}$Marco Pavone is with the Department of Aeronautics and Astronautics, Stanford University, and with NVIDIA
        {\tt\small \{pavone@stanford.edu, mpavone@nvidia.com\}}}%
}

\begin{document}

\maketitle
\thispagestyle{empty}
\pagestyle{empty}

\begin{abstract}

Uncertainty pervades through the modern robotic autonomy stack, with nearly every component (e.g., sensors, detection, classification, tracking, behavior prediction) producing continuous or discrete probabilistic distributions. Trajectory forecasting, in particular, is surrounded by uncertainty as its inputs are produced by (noisy) upstream perception and its outputs are predictions that are often probabilistic for use in downstream planning. However, most trajectory forecasting methods do not account for upstream uncertainty, instead taking only the most-likely values. As a result, perceptual uncertainties are not propagated through forecasting and predictions are frequently overconfident. To address this, we present a novel method for incorporating perceptual state uncertainty in trajectory forecasting, a key component of which is a new statistical distance-based loss function which encourages predicting uncertainties that better match upstream perception. We evaluate our approach both in illustrative simulations and on large-scale, real-world data, demonstrating its efficacy in propagating perceptual state uncertainty through prediction and producing more calibrated predictions.

\end{abstract}

\section{INTRODUCTION}
\begin{figure}[t]
    \centering
    \includegraphics[width=0.85\linewidth]{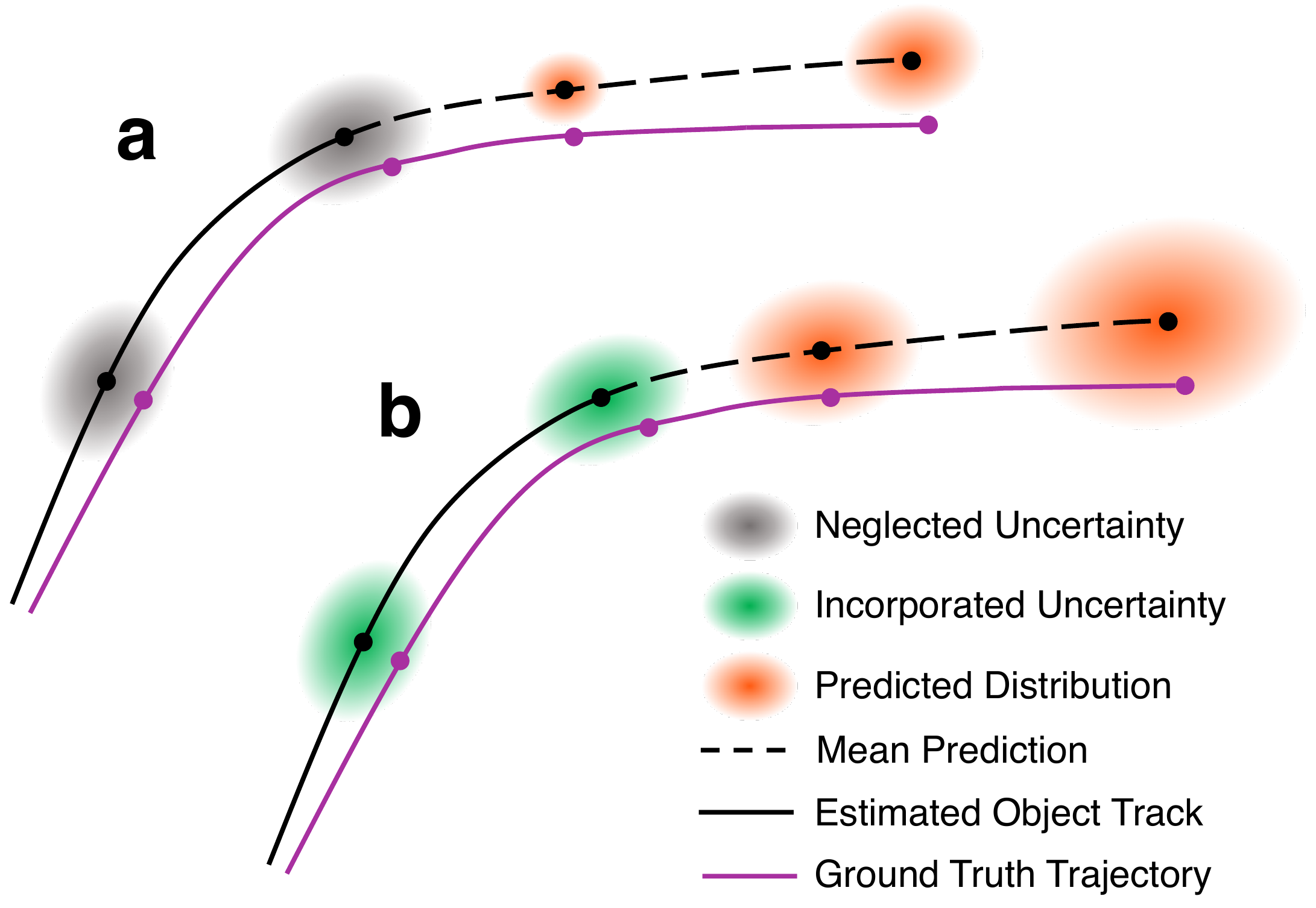}
    
    \vspace*{-0.25cm}
    
    \caption{\textbf{(a)} Most trajectory forecasting methods neglect upstream perceptual uncertainty, e.g., uncertainty in state estimates, assuming that their inputs are known with certainty. As a result, predicted distributions are frequently overconfident. \textbf{(b)} Our method explicitly incorporates and propagates upstream uncertainty to its predictions, yielding more calibrated outputs.}
    \label{fig:problem}
    
    \vspace*{-0.5cm}
    
\end{figure}

Reasoning about perceptual uncertainty and its propagation through the autonomy stack is critical for the safe operation of autonomous vehicles. Failing to do so has unfortunately led to fatalities partially caused by perceptual errors propagated from vision~\cite{teslaCrashReport} and LIDAR-based~\cite{uberCrashReport} systems. 
Currently, however, most trajectory forecasting approaches do not account for upstream uncertainty~\cite{RudenkoPalmieriEtAl2019}, instead taking only the most-likely value (e.g., mean) as input, neglecting measures of uncertainty (e.g., variance). As a result, perceptual uncertainties are not propagated and predictions are frequently overconfident, as shown in \cref{fig:problem}. Such overconfidence is dangerous as an autonomous vehicle may believe it has more free space than in reality, increasing the risk of collisions. A safer approach is to propagate perceptual uncertainty through forecasting systems, enabling planning components to make uncertainty-aware decisions~\cite{mcallister2017concrete,bhatt2020probabilistic}.

Perception systems in autonomous driving are typically comprised of detectors and trackers~\cite{GMSafety2018,UberATGSafety2020,LyftSafety2020,WaymoSafety2021,ArgoSafety2021,MotionalSafety2021,ZooxSafety2021,NVIDIASafety2021}. Broadly, detectors are responsible for identifying objects of interest from raw sensor data
and trackers associate detections of the same object across different timesteps.
Many trackers produce estimates of their uncertainty~\cite{LuoXingEtAl2021}, however they have not yet been incorporated in trajectory forecasting~\cite{RudenkoPalmieriEtAl2019}. 
In the following, we provide an overview of existing approaches for trajectory forecasting and discuss their consideration of perceptual uncertainty.

{\bf Modular Trajectory Forecasting.} Modular autonomy stacks decompose autonomous driving into distinct sub-problems, typically perception, prediction, planning and control~\cite{schwarting2018planning}. This allows for each sub-task to be solved separately and combined through specified interfaces. A typical interface between perception and trajectory forecasting is to only communicate the most likely state estimate for each object detected and tracked by the perception system. Trajectory forecasting methods have thus traditionally assumed their inputs are known with certainty~\cite{lefevre2014survey}. In reality, sensors are imperfect and incorrect assumptions of certainty-equivalence in perception have been partially responsible for two separate fatalities~\cite{teslaCrashReport,uberCrashReport}.

To the best of our knowledge, prior forecasting work has not yet considered the explicit propagation of state uncertainty through modular systems, but there have been many developments~\cite{RudenkoPalmieriEtAl2019}.
For instance, since forecasting is an inherently multimodal task (i.e., there are many possible future outcomes), several works have proposed multimodal probabilistic models, trained using exact-likelihood \cite{rhinehart2018r2p2,chai2019multipath} or variational inference \cite{SchmerlingLeungEtAl2018,IvanovicSchmerlingEtAl2018,IvanovicPavone2019,SalzmannIvanovicEtAl2020}. 
Generative Adversarial Networks (GANs) \cite{GoodfellowPouget-AbadieEtAl2014} can generate empirical trajectory distributions by sampling multiple predictions \cite{GuptaJohnsonEtAl2018,roy2019vehicle}. However, analytic distributions are often useful for gradient-based planning that minimize the likelihood of collisions \cite{schwarting2018planning}. As a result, we focus on methods that predict analytic trajectory distributions. 

{\bf End-to-End Approaches.} End-to-end prediction methods operate directly from raw sensor data, performing detection, tracking, and prediction jointly. FaF~\cite{luo2018fast} introduced the approach of projecting LiDAR points into a bird's eye view (BEV) grid, generating predictions by inferring detections multiple timesteps in the future. This approach was extended by IntentNet~\cite{casas2018intentnet}, which incorporated HD map information as an input, and predicted agent intent as well. SpAGNN~\cite{casas2020spagnn} modeled agent interactions using a graph neural network (GNN), and ILVM~\cite{casas2020implicit} extended this direction further by modeling the joint distribution over future trajectories using a latent variable model. PTP~\cite{WengYuanEtAl2021} also uses a GNN, performing both tracking and prediction in parallel.
These methods only incorporate state uncertainty implicitly, however, making it difficult to transparently analyze, probe (e.g., via counterfactual ``what-if" analyses), and understand the effects of perceptual uncertainty on the rest of the autonomy stack.

{\bf Uncertainty Propagation in Learning.} Existing approaches for uncertainty propagation in machine learning typically view inputs as noisy samples of an underlying data distribution, applying Bayesian neural networks~\cite{Wright1999,WangShiEtAl2016} and Markov models~\cite{AstudilloNeto2011} to estimate the true input distribution and propagate its uncertainty to the output. Our work differs in that it does not need to perform estimation; upstream detectors and trackers typically already characterize their output confidence, e.g., as a Gaussian distribution over states, providing it for use in downstream modules.

Recently, there have been significant efforts on estimating uncertainty in deep learning \cite{blundell2015weight,lakshminarayanan2016simple,gal2017concrete}, especially so in the context of planning \cite{depeweg2016learning,chua2018deep}. However, these works mainly focus on output uncertainty \emph{estimation} rather than input uncertainty \emph{propagation}. Our work tackles the latter.

With such a plethora of available trajectory forecasting approaches, an immediate approach for incorporating state uncertainty might be to augment a model's input with uncertainty information (e.g., concatenating the input's variance). However, as we will show, such an approach is not sufficient.

{\bf Contributions.} Our key contributions are threefold: First, we show that there are structural pitfalls in the standard training pipeline of generative trajectory forecasting methods that hinder the simple incorporation and propagation of state uncertainty (e.g., just adding uncertainty as an input).
Based on this insight, we propose a new training loss that incorporates uncertainty in trajectory forecasting with minimal effects on prediction accuracy (sometimes even improving it).
Finally, we ground our theoretical hypothesis with extensive experiments on illustrative scenarios as well as real-world data.

\section{PROBLEM FORMULATION}

In this work, we aim to generate future trajectory distributions for a time-varying number $N(t)$ of diverse interacting agents $A_1,\dots,A_{N(t)}$, each of which has a semantic type $C_i$ (e.g., Car, Bicycle, Pedestrian). At time $t$, agent $A_i$'s dynamic state $\mathbf{s}_i^{(t)} \in \mathbb{R}^D$ (e.g., position, velocity, orientation) is noisily estimated by an upstream perception system as $\hat{\mathbf{s}}_i^{(t)} \in \mathbb{R}^D$. We assume that the perception system also produces Gaussian state uncertainty information (i.e., state uncertainty covariance $\hat{\mathbf{\Sigma}}_i^{(t)} \in \mathbb{S}_+^{D \times D}$),
with the knowledge that 
many multi-object trackers are based on recursive Bayesian filters~\cite{LuoXingEtAl2021}, which produce such information.

At time $t$, given the estimated state $\hat{\mathbf{s}}_i^{(t)}$, 
associated uncertainty $\hat{\mathbf{\Sigma}}_i^{(t)}$, and their histories 
$\hat{\mathbf{x}}^{(t)} = \hat{\mathbf{s}}_{1,\dots,N(t)}^{(t - H : t)} \in \mathbb{R}^{(H + 1) \times N(t) \times D}$ and $\hat{\mathbf{\Sigma}}^{(t)} = \hat{\mathbf{\Sigma}}_{1,\dots,N(t)}^{(t - H : t)} \in \mathbb{R}^{(H + 1) \times N(t) \times D \times D}$ for the previous $H$ timesteps for each agent, our goal is to produce an accurate and calibrated distribution over all agents' future states for the next $T$ timesteps, $\mathbf{y}^{(t)} = \mathbf{s}_{1,\dots,N(t)}^{(t + 1 : t + T)} \in \mathbb{R}^{T \times N(t) \times D}$, which we denote as $p(\mathbf{y}^{(t)} \mid \hat{\mathbf{x}}^{(t)}, \hat{\mathbf{\Sigma}}^{(t)})$. Note that the output distribution is conditioned on $\hat{\mathbf{\Sigma}}^{(t)}$, differing from prior work. We drop the time superscripts in the rest of the paper for brevity.

\begin{figure*}[t]
    \centering
    \includegraphics[width=0.94\linewidth]{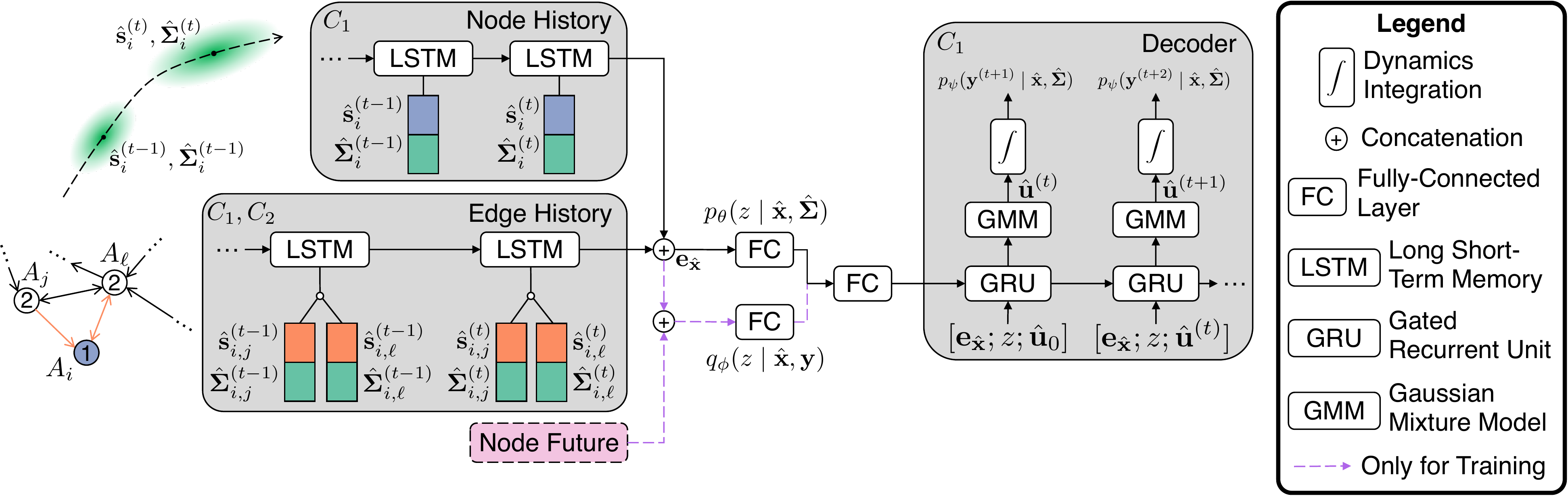}
    
    \vspace*{-0.25cm}
    
    \caption{Our approach represents a scene as a directed spatiotemporal graph where nodes and edges represent agents and their interactions, respectively. Our method incorporates upstream perceptual state uncertainty by encoding state uncertainty information alongside the agent's state. Our novel statistical distance-based loss function term further encourages the incorporation of uncertainty by penalizing overconfident predictions.}
    \label{fig:architecture}
    
    \vspace*{-0.65cm}
    
\end{figure*}

\section{PROPAGATING STATE UNCERTAINTY THROUGH TRAJECTORY FORECASTING}
\label{sec:method}

Our method\footnote{Code online at \url{https://github.com/StanfordASL/PSU-TF}.} for obtaining the desired $p(\mathbf{y} \mid \hat{\mathbf{x}}, \hat{\mathbf{\Sigma}})$ distribution builds upon Trajectron++ \cite{SalzmannIvanovicEtAl2020}, a state-of-the-art multi-agent trajectory forecasting method with publicly-available code. At a high-level, our model (visualized in \cref{fig:architecture}) is a graph-structured recurrent encoder-decoder which makes use of a discrete latent space to explicitly model multimodality. In this section, we describe the core components of the algorithm and highlight our contributions that enable the incorporation and propagation of state uncertainty.

{\bf Input Representation.} We first abstract the scene as a directed spatiotemporal graph $G = (V, E)$, where nodes represent agents and edges represent their directed interactions (allowing for asymmetric influence). As in prior work \cite{AlahiGoelEtAl2016,SalzmannIvanovicEtAl2020}, we use the $\ell_2$ distance as a proxy for agent interaction: an edge connects $A_i$ and $A_j$ if $\| \mathbf{p}_i - \mathbf{p}_j \|_2 \leq d(C_i, C_j)$ where $\mathbf{p}_i, \mathbf{p}_j \in \mathbb{R}^2$ are the 2D positions of agents $A_i, A_j$, respectively, and $d(C_i, C_j)$ is the inter-class distance threshold between agent types $C_i$ and $C_j$ (the order of classes in $d(\cdot, \cdot)$ matters, since we model interactions asymmetrically).

{\bf Encoder.} The encoder models agent history, inter-agent interaction, and any provided scene context (e.g., maps, although they are not used in this work to ensure fair comparisons to prior work in \cref{sec:expts}). In particular, an agent's observed trajectory history (a concatenation of estimated states and their uncertainty) is fed into a Long Short-Term Memory (LSTM) network \cite{HochreiterSchmidhuber1997} with 32 hidden dimensions. 
To model neighboring agents' influence on the modeled agent, we follow \cite{SalzmannIvanovicEtAl2020} and aggregate edge features from neighboring agents with an element-wise sum. 
These aggregated states are then fed into an LSTM with 8 hidden dimensions, yielding a single vector representing the influence that neighboring nodes have on the modeled agent. The node history and edge influence encodings are then concatenated to produce the node representation vector, $\mathbf{e}_{\hat{\mathbf{x}}}$.

{\bf Multimodal Latent Variable.} Our model leverages the Conditional Variational Autoencoder (CVAE) latent variable framework \cite{SohnLeeEtAl2015} to explicitly account for multimodality in future human behavior (i.e., the possibility for many distinct futures). It produces the target $p(\mathbf{y} \mid \hat{\mathbf{x}}, \hat{\mathbf{\Sigma}})$ distribution by introducing a discrete Categorical latent variable $z \in Z$ which encodes high-level latent behavior and allows for the prior distribution $p(\mathbf{y} \mid \hat{\mathbf{x}}, \hat{\mathbf{\Sigma}})$ to be expressed as
\begin{equation}\label{eqn:model}
    p(\mathbf{y} \mid \hat{\mathbf{x}}, \hat{\mathbf{\Sigma}}) = \sum_{z \in Z} p_\psi(\mathbf{y} \mid \hat{\mathbf{x}}, z, \hat{\mathbf{\Sigma}}) p_\theta(z \mid \hat{\mathbf{x}}, \hat{\mathbf{\Sigma}}),
\end{equation}
where $|Z| = 25$ and $\psi, \theta$ are network weights. We chose $|Z|$ as such because it allows for the modeling of a wide variety of high-level latent behaviors and any unused latent classes will be ignored by the CVAE \cite{ItkinaIvanovicEtAl2019}.

{\bf Decoder.} The latent variable $z$ and encoder output $e_\mathbf{x}$ are then passed into the decoder, a 128-dimensional Gated Recurrent Unit (GRU)~\cite{ChoMerrienboerEtAl2014}. Each GRU cell outputs the parameters of a multivariate Normal distribution over controls $\hat{\mathbf{u}}^{(t)}$ (e.g., acceleration and steering rate).
The agent's linear(ized) system dynamics are then integrated with the forecasted controls $\hat{\mathbf{u}}^{(t)}$ to obtain trajectories in position space~\cite{SalzmannIvanovicEtAl2020}.
Importantly, the only source of uncertainty in the agent's dynamics is the decoder's output. Thus, the decoder's output uncertainty can be directly propagated to position space~\cite{Kalman1960,ThrunBurgardEtAl2005EKF}. Predicting agent controls and integrating them through dynamics has been shown to improve forecasting accuracy and ensures that the predictions are dynamically feasible~\cite{SalzmannIvanovicEtAl2020}.

{\bf Specifying the Loss Function.} We base our model's loss function on the discrete InfoVAE \cite{ZhaoSongEtAl2019} objective used for Trajectron++~\cite{SalzmannIvanovicEtAl2020}. To ensure that the model makes use of the encoded state uncertainty, we introduce an additional term that penalizes the statistical distance between the predicted distribution and the tracked GT data. The specific choice of statistical distance $D_{SD}$ will be discussed in \cref{sec:method}. Formally, we train the neural network weights $\phi, \theta, \psi$ to maximize the following for each agent $A_i$,
\begin{equation}\label{eqn:loss_fn}
\begin{aligned}
&\mathbb{E}_{z \sim q_\phi(\cdot \mid \hat{\mathbf{x}}_i, \mathbf{y}_i)} \big[\log p_\psi(\mathbf{y}_i \mid \hat{\mathbf{x}}_i, z, \textcolor{red}{\hat{\mathbf{\Sigma}}_i})\\
&\, - \textcolor{blue}{D_{SD}(p_\psi(\mathbf{y}_i \mid \hat{\mathbf{x}}_i, z, \hat{\mathbf{\Sigma}}_i) \parallel \mathcal{N}(\mathbf{y}_i, \hat{\mathbf{\Sigma}}_i))} \big]\\
&\, - \beta D_{KL}\big(q_\phi(z \mid \hat{\mathbf{x}}_i, \mathbf{y}_i) \parallel p_\theta(z \mid \hat{\mathbf{x}}_i, \textcolor{red}{\hat{\mathbf{\Sigma}}_i})\big) + I_{q}(\hat{\mathbf{x}}_i; z),
\end{aligned}
\end{equation}
where $I_q$ is the mutual information between $\hat{\mathbf{x}}_i$ and $z$ under the distribution $q_\phi(\hat{\mathbf{x}}_i,z)$ and $D_{SD}$ is a measure of the distance between two probability distributions.
During training, a bi-directional LSTM with 32 units encodes a node's GT future trajectory, producing $q_\phi(z \mid \mathbf{x}, \mathbf{y})$~\cite{SohnLeeEtAl2015}.

To summarize, our method differs from Trajectron++ \cite{SalzmannIvanovicEtAl2020} by its introduction of state uncertainty information as an input (in \textcolor{red}{red}) and the addition of a statistical distance loss term (in \textcolor{blue}{blue}). In the remainder of this section, we describe why both of these components are necessary.

Introducing state uncertainty information as an input is important as it conditions our model's predictions on the existing level of uncertainty.
However, only adding state uncertainty information does not yield changes in the model's output uncertainty because the log-probability term in \cref{eqn:loss_fn} encourages overconfidence (shown in \cref{sec:expts}). In particular, it is maximized when the output distribution
lies directly on top of the GT future position with an infinitesimal uncertainty (a Dirac delta distribution). This trend towards overconfidence is a general downside of training by maximizing log-probability. Adding a statistical distance term balances the loss function by encouraging the predicted distribution to be closer to the GT tracked distribution, and thus also more calibrated (shown in \cref{sec:expts}) since statistical distance is minimized when the predicted and GT distributions are the same.

{\bf Choice of Statistical Distance.}
There are many options for the statistical distance $D_{SD}$ in \cref{eqn:loss_fn}. To be practical, $D_{SD}$ must be: (1) a measure of distance between distributions, (2) differentiable over the space of distributions, 
and (3) efficient to compute. The first consideration is required as $D_{SD}$ computes the distance between distributions in \cref{eqn:loss_fn} and the second is necessary from a learning perspective.
Finally, efficiency is necessary as $D_{SD}$ will be computed many times during training. 
While not a core desideratum, we also found that a symmetric $D_{SD}$ is desirable because over- and underestimates are then viewed as equally inaccurate.
Asymmetrical evaluation has been studied in prediction \cite{CasasGulinoEtAl2020,IvanovicPavone2021}, and can also be included in this work via asymmetric statistical distance measures. However, we leave the decision to include asymmetry to practitioners.

To choose $D_{SD}$, we implemented common statistical distances that satisfy the above desiderata and have closed-form expressions when evaluated between Gaussians, namely the Symmetric Kullback-Leibler (SKL), Hellinger (He), and Bhattacharyya (Bh) distances. To understand each measure's behavior, we  
computed distances between manually-specified pairs of Gaussian distributions, varying their means $\mu_P, \mu_Q$ and covariances $\Sigma_P, \Sigma_Q$.
We found that He frequently saturates at its maximum value of $1$ (which can stall training).
SKL
and Bh yielded similar values, but Bh was faster to compute.
Thus, we use the Bh distance for $D_{SD}$ in \cref{eqn:loss_fn}. In particular, we implement an extension that computes the distance between a Gaussian Mixture Model~(GMM) $P$ with $K$ components and a Gaussian $Q$~\cite{SfikasConstantinopoulosEtAl2005}:
\begin{equation}\label{eqn:bh_gmm}
\begin{aligned}
D_{SD}(P, Q) &= \sum_{k=1}^K \pi_k\, D_B(\mathcal{N}(\mu_k, \Sigma_k), \mathcal{N}(\mu_Q, \Sigma_Q))
\end{aligned}
\end{equation}
where 
$D_B(\cdot, \cdot)$ is the closed-form Bh distance between two unimodal Gaussians~\cite{Fukunaga1990} and $P = \sum_{k=1}^K \pi_k\, \mathcal{N}(\mu_k, \Sigma_k)$.

\section{EXPERIMENTS AND ANALYSES} \label{sec:expts}
Our method is evaluated on an illustrative charged particle system and three publicly-available datasets: The ETH \cite{PellegriniEssEtAl2009}, UCY \cite{LernerChrysanthouEtAl2007}, and nuScenes \cite{CaesarBankitiEtAl2019} datasets. We also implement a detection-tracking perception system from state-of-the-art components and show the performance of our approach in a practical perception-prediction stack, using raw nuScenes data for sensor observations.
The charged particle system serves as a controlled experiment and demonstrates that our approach indeed takes state uncertainty into account when generating predictions. The ETH, UCY, and nuScenes datasets evaluate our approach's ability to model real-world pedestrians and vehicles.

Our model was implemented with PyTorch on a desktop computer running Ubuntu 18.04 containing an AMD Ryzen 1800X CPU and two NVIDIA GTX 1080 Ti GPUs. We trained the model for 100 epochs on the particle and pedestrian datasets and 20 epochs on the nuScenes dataset.

{\bf Baselines.} 
We compare our work to Trajecton++ \cite{SalzmannIvanovicEtAl2020} (``T++"), whose loss function only maximizes the log-probability of the GT under the predicted distribution. We also compare against a model that only minimizes the statistical distance function $D_{SD}$ (``$D_{SD}$ Only"). Other domain-specific baselines are introduced in their respective sections.

{\bf Methodology.} 
For the three real-world datasets, GT state uncertainty is obtained by running an Extended Kalman Filter (EKF) on vehicles (modeled as bicycles \cite{PadenCapEtAl2016}), and a Kalman Filter (KF) on pedestrians (modeled as single integrators). All filter covariances are initialized as identity.
Since the state uncertainties are estimated and not provided by the datasets, we do \emph{not} use them in our evaluation. Instead, we evaluate our model with the following metrics:
\begin{enumerate}
    \item Negative Log Likelihood (NLL): Mean NLL of the GT trajectory under a distribution generated by the model.
    \item Final Displacement Error (FDE): $\ell_2$ distance between the predicted mean final position and the GT final position at the prediction horizon $T$.
    \item Delta Empirical Sigma Value ($\Delta\text{ESV}_i$) \cite{PostnikovGamayunovEtAl2021}: The difference in the fraction of GT positions that fall within the $i$-$\sigma$ level set (e.g., 1$\sigma$, 2$\sigma$, 3$\sigma$) of the predicted distribution and the fraction from an ideal
    Gaussian.
\end{enumerate}
In particular, $\Delta\text{ESV}_i$ is a useful metric for identifying over- or underconfidence, as $\Delta\text{ESV}_i := \sigma_{\text{pred},i} - \sigma_{\text{ideal},i}$ where $\sigma_{\text{pred},i}$ is the empirical fraction of GT positions that lie within the $i$-sigma level set of the prediction distribution and $\sigma_{\text{ideal},i}$ is the expected fraction from a perfectly-calibrated bivariate Gaussian, where $\sigma_{\text{ideal},1} \approx 0.39, \sigma_{\text{ideal},2} \approx 0.86,$ and $\sigma_{\text{ideal},3} \approx 0.99$~\cite{Bajorski2011}. Thus, $\Delta\text{ESV}_i < 0$ indicates overconfidence and $\Delta\text{ESV}_i > 0$ signifies underconfidence.
    
\begin{table}[t]
    \centering
    \setlength\tabcolsep{4.5pt}
    \caption{Bolded is best, underlined is second-best.}
    
    \vspace*{-0.25cm}
    
    \begin{tabular}{l|cccc}
    \toprule
    \multicolumn{1}{c|}{\textbf{Particles}} & \multicolumn{4}{c}{\textbf{NLL (nats)}}\\ \cline{1-5}
    \multicolumn{1}{c|}{\textbf{Horizon}} & 0.2s & 0.4s & 0.6s & 0.8s \Tstrut \\
    \midrule 
    T++ \cite{SalzmannIvanovicEtAl2020} & 
    \textbf{-5.70}\tiny{$\pm0.96$} & 
    \textbf{-5.29}\tiny{$\pm1.41$} & 
    \textbf{-4.68}\tiny{$\pm1.81$} & 
    \textbf{-4.03}\tiny{$\pm2.10$}\\

    $D_{SD}$ Only & 
    0.44\tiny{$\pm0.13$}& 
    0.47\tiny{$\pm0.25$}& 
    0.54\tiny{$\pm0.50$}& 
    0.67\tiny{$\pm0.78$}\\
    \hline

    Ours & 
    \underline{-5.53}\tiny{$\pm1.03$}&
    \underline{-4.92}\tiny{$\pm1.40$}&
    \underline{-4.12}\tiny{$\pm1.65$}&
    \underline{-3.37}\tiny{$\pm1.82$}\Tstrut \\
    \end{tabular}

    \setlength\tabcolsep{1.5pt}
    \begin{tabular}{l|c|cccc|cccc}
    \toprule
    \multicolumn{2}{c|}{} & \multicolumn{4}{c|}{\textbf{$\Delta$ESV$_i$}} & \multicolumn{4}{c}{\textbf{FDE (m)}} \\ \cline{1-10}
    \multicolumn{2}{c|}{\textbf{Horizon}} & 0.2s & 0.4s & 0.6s & 0.8s & 0.2s & 0.4s & 0.6s & 0.8s \Tstrut \\
    \midrule 
    \multirow{3}{*}{T++ \cite{SalzmannIvanovicEtAl2020}}
    & $\Delta\text{ESV}_1$ & \textbf{-0.36} & \textbf{-0.32} & \textbf{-0.31} & \textbf{-0.31} \\
    & $\Delta\text{ESV}_2$ & \textbf{0.04} & \textbf{-0.03} & \underline{-0.11} & -0.17
    & \textbf{0.03} & \textbf{0.08} & \textbf{0.20} & \textbf{0.40}\\
    & $\Delta\text{ESV}_3$ & \underline{-0.01} & -0.04 & -0.09 & -0.12 \\
    \cline{2-10}

    \multirow{3}{*}{$D_{SD}$ Only}
    & $\Delta\text{ESV}_1$ & 0.61 & 0.60 & 0.58 & 0.55 \Tstrut \\
    & $\Delta\text{ESV}_2$ & 0.14 & 0.14 & 0.13 & \underline{0.12}
    & 0.06 & 0.10 & 0.22 & 0.42\\
    & $\Delta\text{ESV}_3$ & \underline{0.01} & \underline{0.01} & \textbf{7e-3} & \textbf{-2e-3} \\
    \midrule 

    \multirow{3}{*}{Ours}
    & $\Delta\text{ESV}_1$ & \underline{0.57} & \underline{0.51} & \underline{0.44} & \underline{0.38} \\
    & $\Delta\text{ESV}_2$ & \underline{0.13} & \underline{0.11} & \textbf{0.08} & \textbf{0.05} 
    & \underline{0.04} & \underline{0.09} & \underline{0.21} & \underline{0.41}\\
    & $\Delta\text{ESV}_3$ & \textbf{8e-3} & \textbf{-2e-3} & \underline{-0.01} & \underline{-0.03} \\
    \bottomrule
    \end{tabular}
    \label{tab:particles_quant}
    
    \vspace*{-0.5cm}
    
\end{table}

\subsection{Illustrative Charged-Particle Simulation}
In this evaluation, we replace complex interactions between real-world agents with well-understood, controlled dynamics and verify that our method is able to incorporate state uncertainty and propagate such information to its output.

\textbf{Dataset.}
We simulate a charged particle system with 3 agents, where the particles (agents) are modeled as double integrators and interact according to the Social Forces Model \cite{HelbingMolnar1995}. A state uncertainty $\Sigma$ is generated for each particle agent, where the variance in the $x$ and $y$ directions is sampled from a Gaussian distribution. Additional Gaussian noise is added to the variances at each timestep of the trajectory to simulate noise produced by an actual perception system.
We collect train, validation, and test sets by initializing agents with random positions and velocities and simulating 250, 75, and 50 randomized scenarios, respectively, for 30s at 10Hz.

\textbf{Quantitative Results.}
\cref{tab:particles_quant} shows that, as expected, T++ has the best NLL and FDE values across all prediction horizons. This makes sense because its loss function solely optimizes NLL and does not account for state uncertainty. As a result, T++ produces overconfident distributions
(evidenced by its negative $\Delta\text{ESV}_i$ values).

In comparison, the $D_{SD}$ Only model has the worst NLL and FDE values, but the best (lowest) NLL standard deviation and highest $\Delta\text{ESV}_i$ values. This makes sense as the model focuses only on matching its output distribution to the estimated GT state uncertainty, resulting in large positive $\Delta\text{ESV}_i$ values as $D_{SD}$ Only produces underconfident distributions to try and cover the estimated GT distribution.

Our approach combines the best of both worlds,
achieving the best or second-best results across all metrics.
As mentioned in \cref{sec:method}, NLL is minimized when Dirac delta functions are placed on top of GT positions, therefore correctly propagating uncertainty should lead to a reduction in pure-NLL performance (and is why our method performs second-best on NLL). By the same argument, our model performs better than $D_{SD}$ Only on the $\Delta\text{ESV}_i$ metrics because \cref{eqn:loss_fn} balances distributional and prediction accuracy, improving calibration overall.

\begin{figure}[t]
    \centering
    \includegraphics[width=0.48\linewidth, frame]{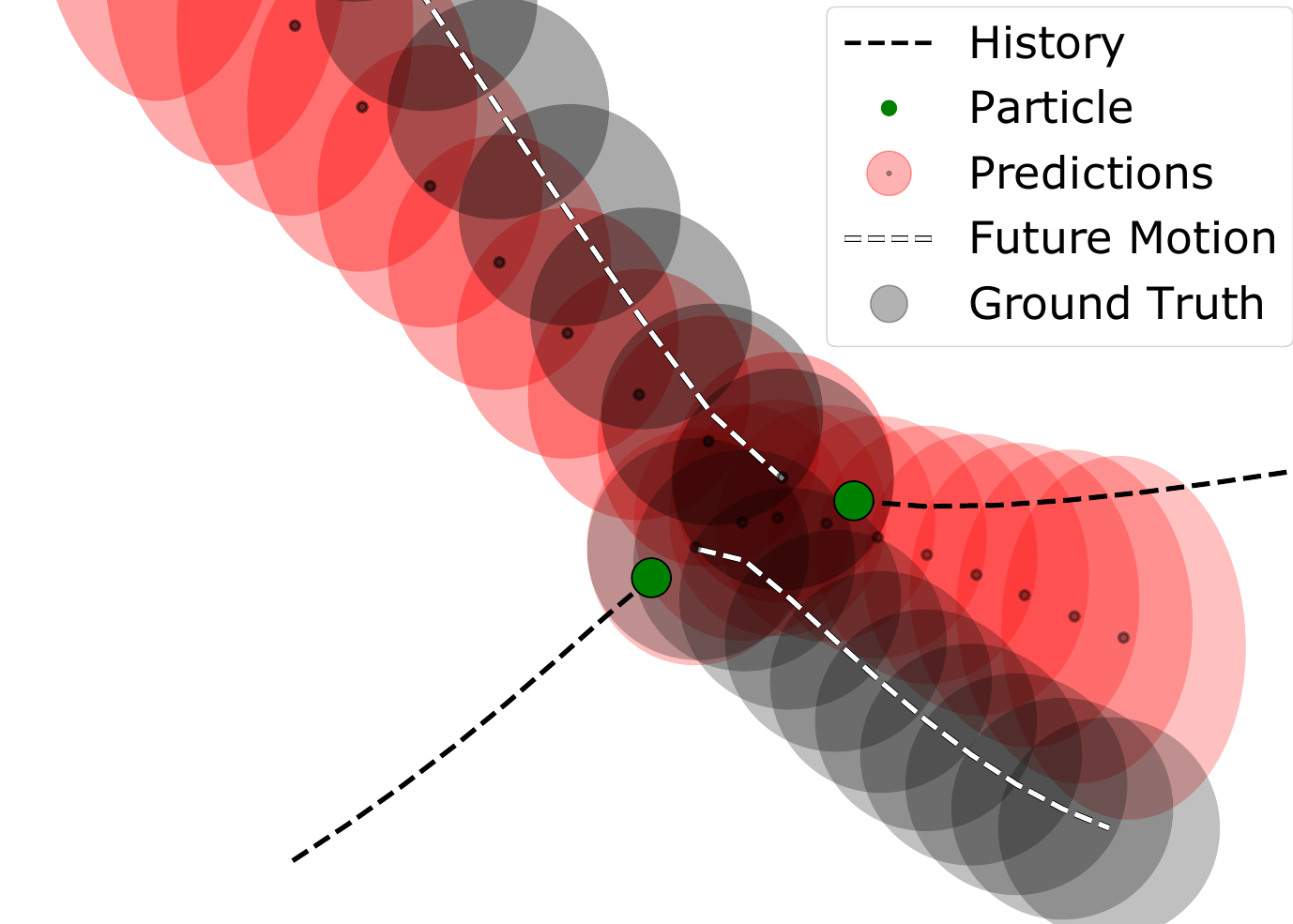}
    \includegraphics[width=0.48\linewidth,frame]{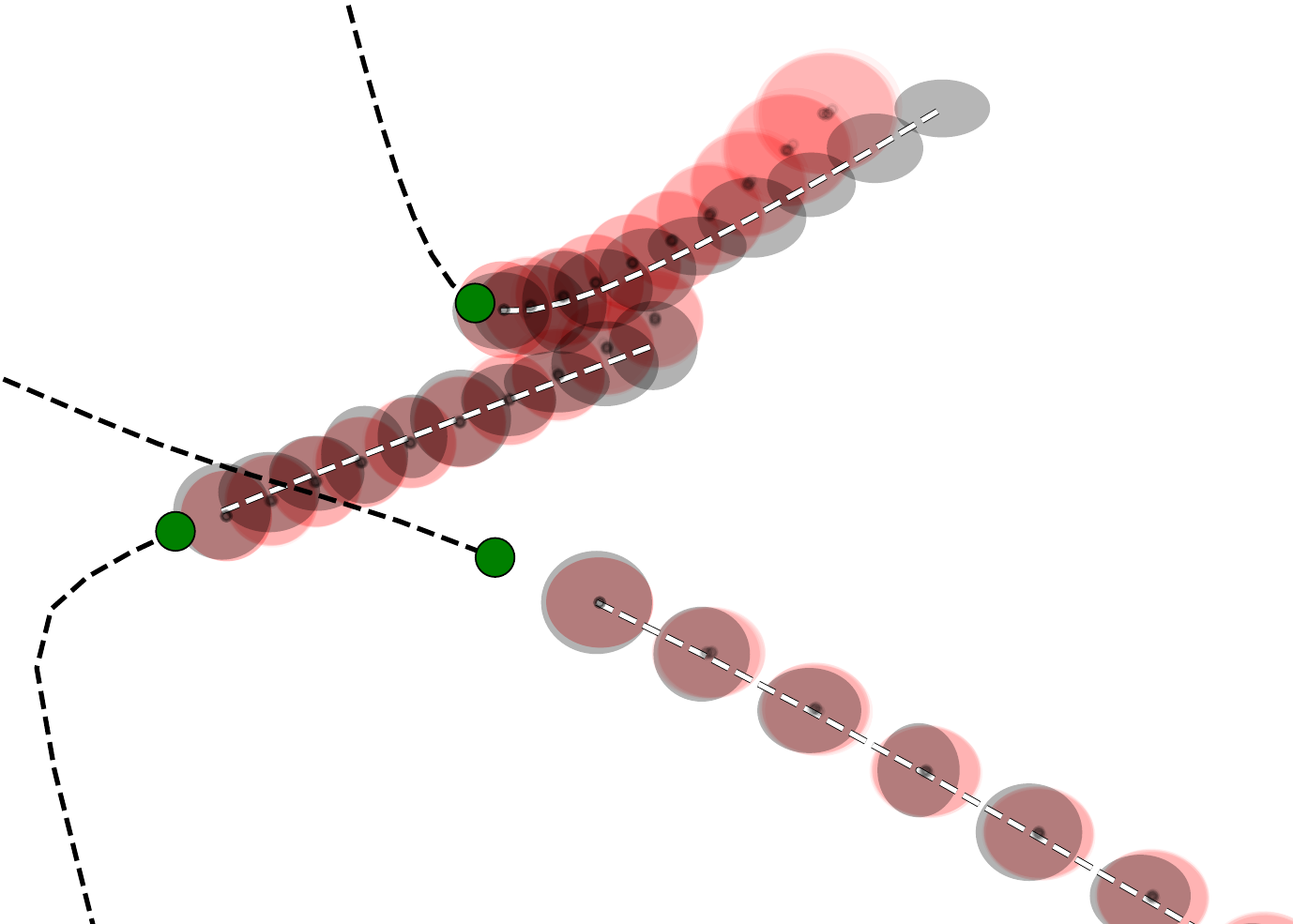}
    \caption{Our method can zero-shot generalize to unseen uncertainty scales. Even if our method is only trained on (manually-specified) large-scale uncertainties~(\textbf{left}), it is still able to propagate input uncertainty and maintain the appropriate scale when evaluated on small-scale uncertainties~(\textbf{right}).}
    \label{fig:particles_gen}

\end{figure}

\textbf{Generalization to Unseen Uncertainty Scales.}
Introducing $D_{SD}$ to \cref{eqn:loss_fn} affords the model zero-shot generalization capabilities to different state uncertainty scales. To verify this, we trained a version of
our model on the particles dataset with only (manually-specified) large GT covariances, testing it on large and small covariance data. In \cref{fig:particles_gen}, we see that, despite the model having only seen large uncertainties during training, it is able to generalize and maintain performance with small-scale uncertainty.

\subsection{Pedestrian Motion Forecasting}

\textbf{Dataset.}
The ETH \cite{PellegriniEssEtAl2009} and UCY \cite{LernerChrysanthouEtAl2007} datasets consist of real pedestrian trajectories with rich multi-human interaction scenarios captured at 2.5 Hz ($\Delta t = 0.4s$). In total, there are 5 sets of data, 4 unique scenes, and 1536 unique pedestrians. They are a standard benchmark in the field, containing challenging behaviors such as couples walking together, groups crossing each other, and groups forming and dispersing. We simulate an upstream perception system by running a KF on the GT pedestrian positions. This provides the state uncertainty $\Sigma$ for training. As in prior work \cite{SalzmannIvanovicEtAl2020}, a leave-one-out strategy is used for evaluation, where the model is trained on four datasets and evaluated on the held-out fifth. An observation length of 12 timesteps (4.8s) is used for evaluation.

\textbf{Quantitative Results.}
\cref{tab:predestrian_quant} summarizes the results averaged across the five pedestrian datasets. In addition to the original baselines, we also compare to prior pedestrian forecasting methods \cite{AlahiGoelEtAl2016,VemulaMuellingEtAl2018} as well as traditional baselines (i.e., Linear and LSTM from \cite{SalzmannIvanovicEtAl2020}). Similar to the particles dataset, we see that T++ performs best on the NLL and FDE metrics but also has the largest NLL variance and worst $\Delta\text{ESV}_i$ values. $D_{SD}$ Only has the lowest NLL variance and largest $\Delta\text{ESV}_i$. Our method still performs best or second best across all metrics except FDE, however the difference in FDE is very small ($\leq$ 0.03m). This shows that our method is able to maintain distributional and mean accuracy on real-world pedestrian data.
    
\begin{table}[t]
    \centering
    \setlength\tabcolsep{3pt}
    \caption{Bolded is best, underlined is second-best.}
    \begin{tabular}{l|cccc}
    \toprule
    \multicolumn{1}{c|}{\textbf{ETH/UCY}} & \multicolumn{4}{c}{\textbf{NLL (nats)}}\\ \cline{1-5}
    \multicolumn{1}{c|}{\textbf{Horizon}} & 1.2s & 2.4s & 3.6s & 4.8s \Tstrut \\
    \midrule 
    T++ \cite{SalzmannIvanovicEtAl2020} & 
    \textbf{-3.97}\tiny{$\pm1.56$} & 
    \textbf{-2.34}\tiny{$\pm2.00$} & 
    \textbf{-1.30}\tiny{$\pm2.21$} & 
    \textbf{-0.54}\tiny{$\pm2.31$}\\

    $D_{SD}$ Only & 
    1.58\tiny{$\pm0.04$}& 
    1.65\tiny{$\pm0.10$}& 
    1.79\tiny{$\pm0.25$}& 
    2.01\tiny{$\pm0.45$}\\
    \hline

    Ours & 
    \underline{-3.43}\tiny{$\pm1.26$}&
    \underline{-1.95}\tiny{$\pm1.53$}&
    \underline{-0.99}\tiny{$\pm1.67$}&
    \underline{-0.27}\tiny{$\pm1.79$}\Tstrut \\
    \end{tabular}

    \setlength\tabcolsep{1.5pt}
    \begin{tabular}{l|c|cccc|cccc}
    \toprule
    \multicolumn{2}{c|}{} & \multicolumn{4}{c|}{\textbf{$\Delta$ESV$_i$}} & \multicolumn{4}{c}{\textbf{FDE (m)}} \\ \cline{1-10}
    \multicolumn{2}{c|}{\textbf{Horizon}} & 1.2s & 2.4s & 3.6s & 4.8s & 1.2s & 2.4s & 3.6s & 4.8s \Tstrut \\
    \midrule 
    \multicolumn{2}{l|}{Linear} & - & - & - & - & - & - & - & 1.59 \\
    \multicolumn{2}{l|}{LSTM} & - & - & - & - & - & - & - & 1.52 \\
    \multicolumn{2}{l|}{S-LSTM \cite{AlahiGoelEtAl2016}} & - & - & - & - & - & - & - & 1.54 \\
    \multicolumn{2}{l|}{S-ATTN \cite{VemulaMuellingEtAl2018}} & - & - & - & - & - & - & - & 2.59 \\
    \hline
    \multirow{3}{*}{T++ \cite{SalzmannIvanovicEtAl2020}}
    & $\Delta\text{ESV}_1$ & \textbf{-0.28} & \underline{-0.30} & \underline{-0.31} & \underline{-0.31} \Tstrut \\
    & $\Delta\text{ESV}_2$ & -0.29 & -0.47 & -0.54 & -0.59 
    & \textbf{0.10} & \textbf{0.37} & \textbf{0.72} & \textbf{1.13}\\
    & $\Delta\text{ESV}_3$ & -0.24 & -0.39 & -0.44 & -0.48 \\
    \cline{2-10}

    \multirow{3}{*}{$D_{SD}$ Only}
    & $\Delta\text{ESV}_1$ & 0.61 & 0.58 & 0.52 & 0.46 \Tstrut \\
    & $\Delta\text{ESV}_2$ & \underline{0.14} & \underline{0.14} & \underline{0.13} & \textbf{0.11} 
    & \underline{0.12} & \underline{0.39} & \underline{0.73} & \textbf{1.13}\\
    & $\Delta\text{ESV}_3$ & \textbf{0.01} & \textbf{0.01} & \textbf{9e-3} & \textbf{4e-3} \\
    \midrule 

    \multirow{3}{*}{Ours}
    & $\Delta\text{ESV}_1$ & \underline{0.36} & \textbf{0.14} & \textbf{0.02} & \textbf{-0.05} \\
    & $\Delta\text{ESV}_2$ & \textbf{0.07} & \textbf{-0.04} & \textbf{-0.12} & \underline{-0.18} 
    & \underline{0.12} & 0.40 & 0.75 & \underline{1.16}\\
    & $\Delta\text{ESV}_3$ & \textbf{-0.01} & \underline{-0.06} & \underline{-0.10} & \underline{-0.14} \\
    \bottomrule
    \end{tabular}
    \label{tab:predestrian_quant}
    
    \vspace{-0.5cm}
    
\end{table}
    
\begin{figure}[t]
    \centering
    \includegraphics[width=0.32\linewidth, frame]{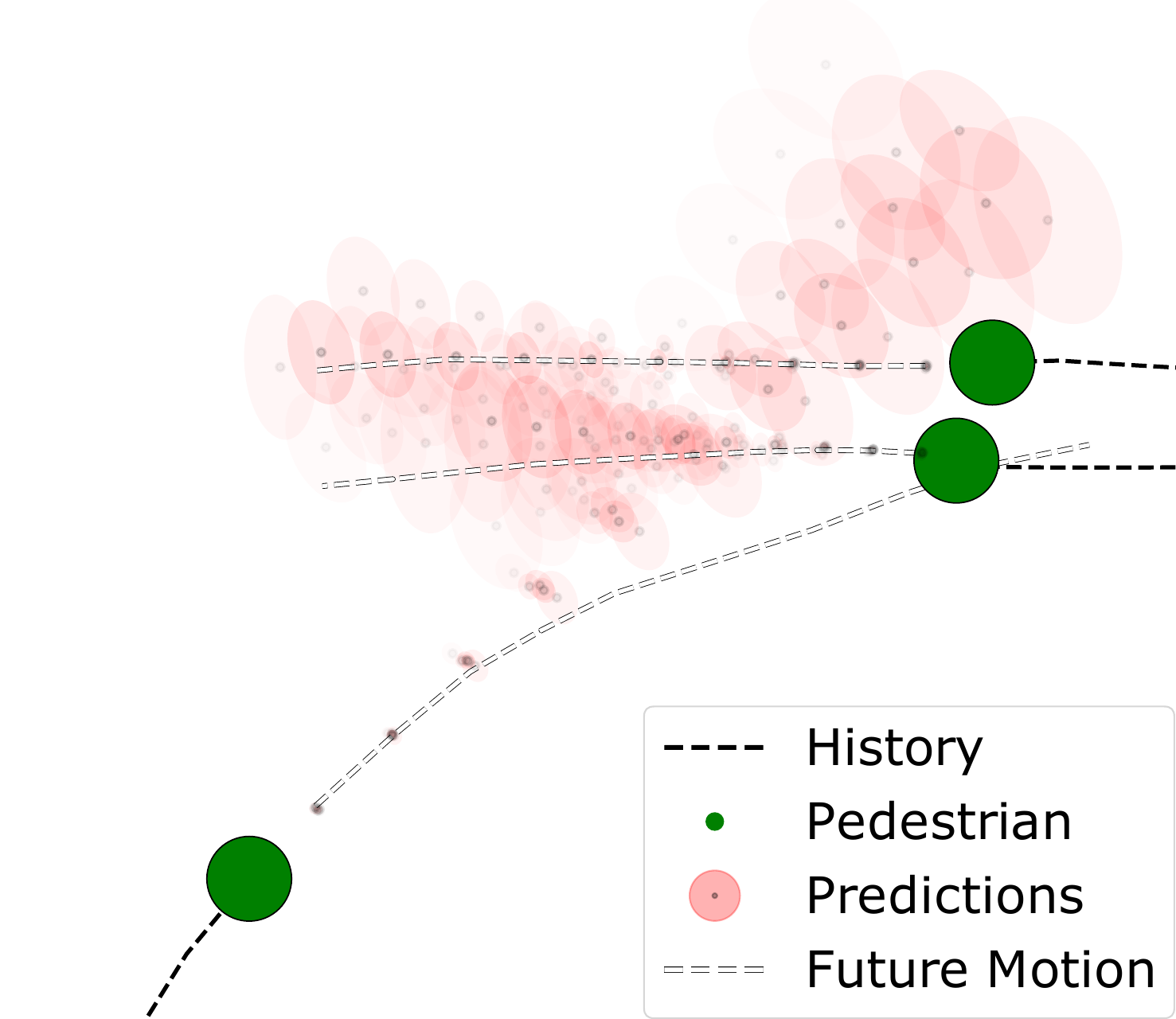}
    \includegraphics[width=0.32\linewidth, frame]{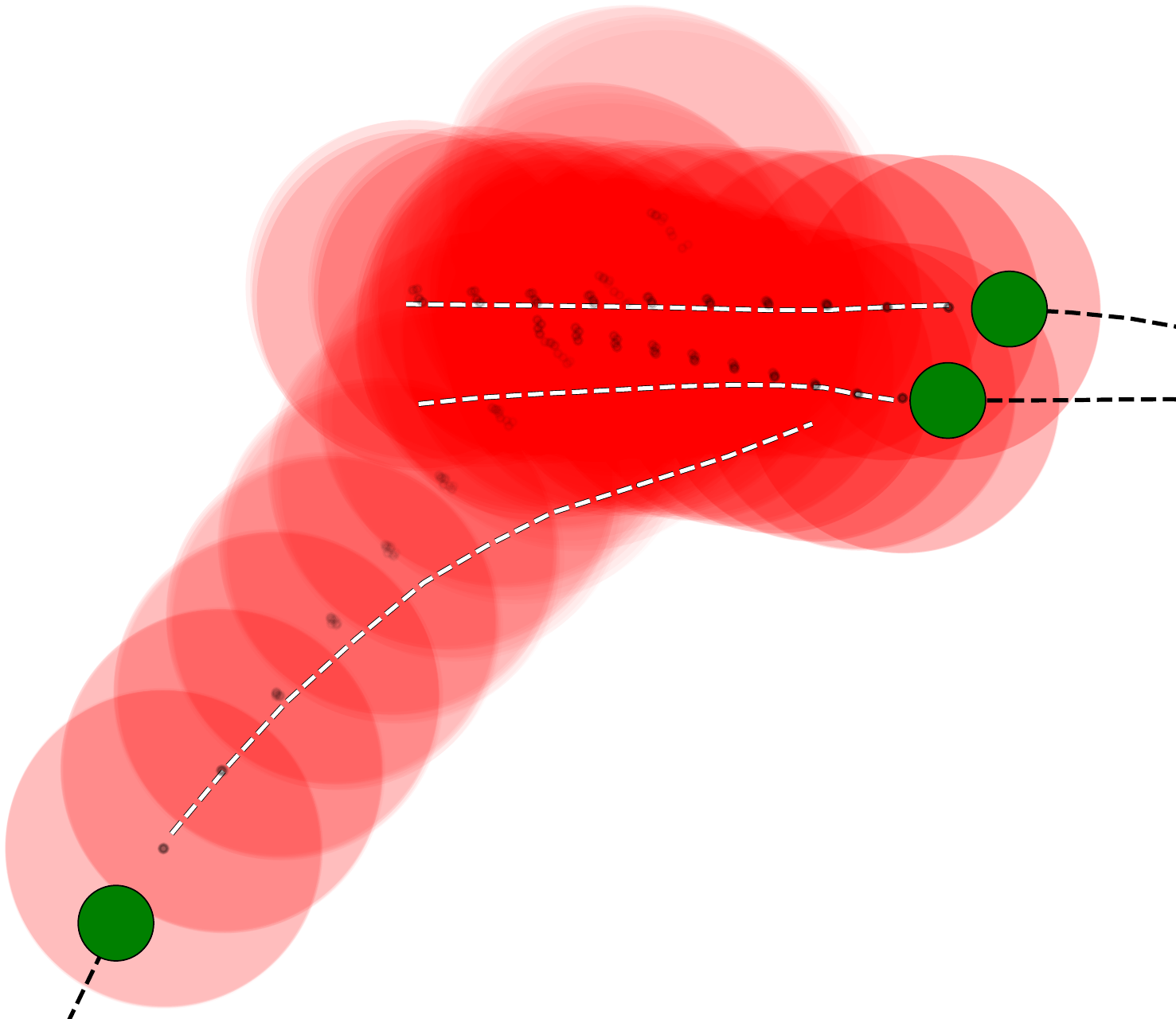}
    \includegraphics[width=0.32\linewidth, frame]{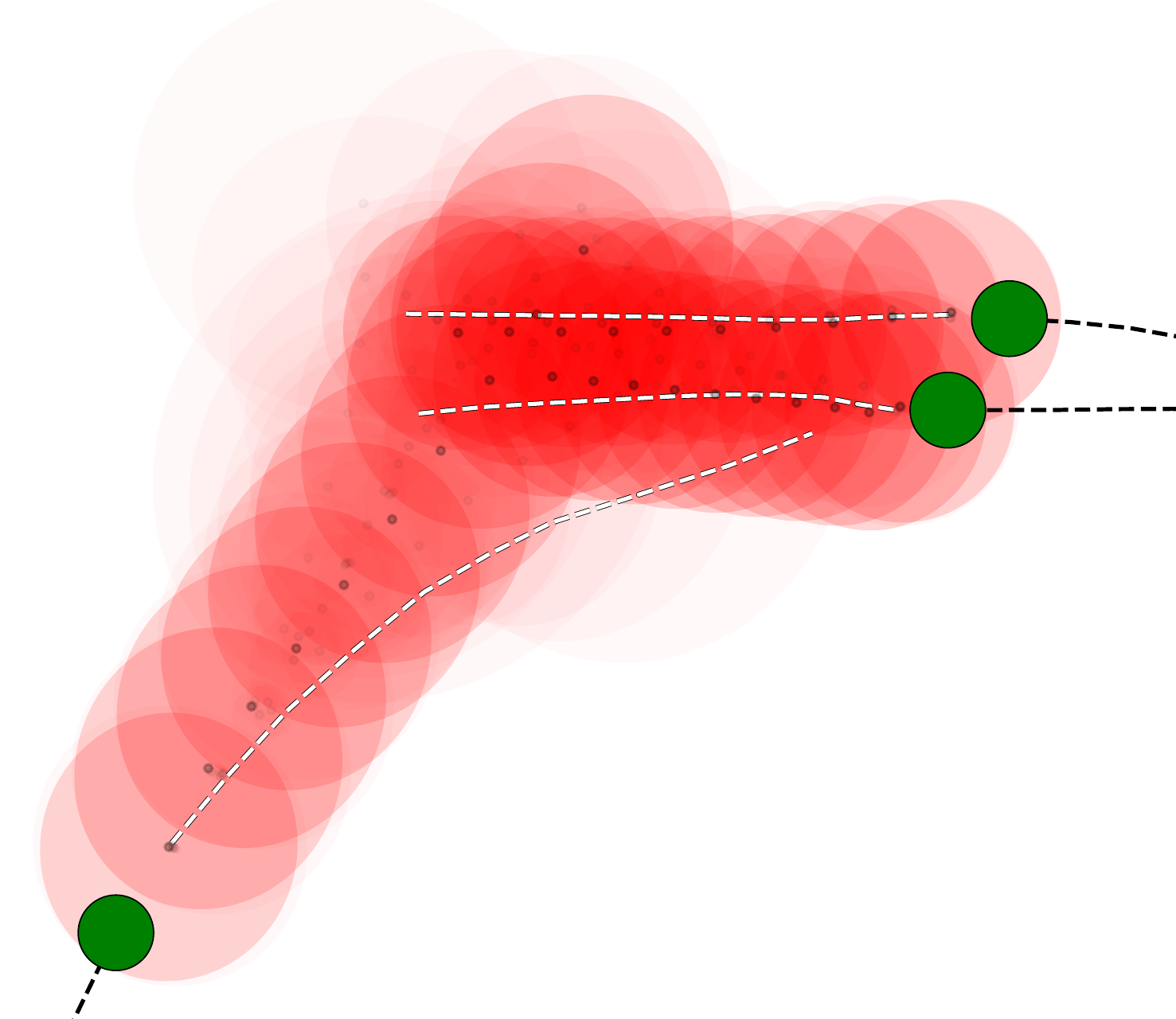}
    \caption{Trajectron++'s predictions (\textbf{left}) are overconfident due to its ignorance of upstream uncertainty. By comparison, $D_{SD}$ Only (\textbf{middle}) and our method (\textbf{right}) produce uncertainties that accurately reflect where the pedestrians might be. Our method's mean predictions (black dots) are more accurate due to the inclusion of the log-probability loss function term.}
    \label{fig:pedestrian_qual}
    
    \vspace{-0.5cm}
    
\end{figure}

\textbf{Qualitative Results.}
We see in \cref{fig:pedestrian_qual} that T++ (still) produces overconfident distributions, especially during the first few timesteps. We also observe that T++ produces several GMM component distributions with roughly equal positions and mixing probabilities. This suggests that T++ uses its latent variable (which produces the GMM mixing probabilities) to model uncertainty, while producing tight distributions to minimize the GT's NLL. $D_{SD}$ Only produces much larger distributions and, as a result, its predictions are unimodal.
Our method produces tighter distributions relative to $D_{SD}$ Only, and some other latent modes are visible, visualized by the faint ellipses around the most-likely prediction (opacity is proportional to mode probability). 

\subsection{Autonomous Driving}

\textbf{Dataset.}
The nuScenes dataset is a large-scale dataset for autonomous driving with 1000 scenes in Boston and Singapore \cite{CaesarBankitiEtAl2019}. Each scene is annotated at 2 Hz ($\Delta t = 0.5s$) and is 20s long, containing up to 23 semantic object classes.
As before, we simulate an upstream perception system by running an EKF on the GT vehicle positions and a KF on the GT pedestrian positions to obtain $\Sigma$ (initialized as $I$). %

\textbf{Quantitative Results.}
\cref{tab:nuscenes_quant} summarizes the model's performance for predicting vehicles. In addition to the original baselines, we also compare to existing works which predict vehicle trajectories from tracked GT detections \cite{GuptaJohnsonEtAl2018,ChandraBhattacharyaEtAl2019,DeoTrivedi2018,ChandraGuanEtAl2020}. We can see that our model is still the best or second-best across all metrics. In this case, our method actually outperforms T++ on FDE for longer horizons. The model also consistently outperforms T++ on the $\Delta\text{ESV}_i$ metrics. These results decidedly show that our work is able to meaningfully propagate state uncertainty and still maintain high prediction accuracy.

\begin{table}[t]
    \centering
    \setlength\tabcolsep{4pt}
    \caption{nuScenes vehicles with GT detections tracked by an EKF. Bolded is best, underlined is second-best.}
    \begin{tabular}{l|ccc}
    \toprule
    \multicolumn{1}{c|}{\textbf{Vehicles}} & \multicolumn{3}{c}{\textbf{NLL (nats)}}\\ \cline{1-4}
    \multicolumn{1}{c|}{\textbf{Horizon}} & 1s & 2s & 3s \Tstrut \\
    \midrule 
    T++ \cite{SalzmannIvanovicEtAl2020} & 
    \textbf{-2.36}\tiny{$\pm3.00$} & 
    \textbf{-1.57}\tiny{$\pm3.20$} & 
    \textbf{-0.88}\tiny{$\pm3.43$}\\

    $D_{SD}$ Only & 
    1.96\tiny{$\pm0.25$}& 
    2.32\tiny{$\pm0.54$}& 
    2.70\tiny{$\pm0.92$}\\
    \hline

    Ours & 
    \underline{-2.17}\tiny{$\pm2.53$}&
    \underline{-1.36}\tiny{$\pm2.81$}&
    \underline{-0.65}\tiny{$\pm3.04$}\Tstrut \\
    \end{tabular}

    \begin{tabular}{l|c|ccc|ccc}
    \toprule
    \multicolumn{2}{c|}{} & \multicolumn{3}{c|}{\textbf{$\Delta$ESV$_i$}} & \multicolumn{3}{c}{\textbf{FDE (m)}} \\ \cline{1-8}
    \multicolumn{2}{c|}{\textbf{Horizon}} & 1s & 2s & 3s & 1s & 2s & 3s \Tstrut \\
    \midrule 
    \multicolumn{2}{l|}{Conv-Social \cite{DeoTrivedi2018}} & - & - & - & 0.78 & - & 3.02\\
    \multicolumn{2}{l|}{Social GAN \cite{GuptaJohnsonEtAl2018}} & - & - & - & 0.59 & - & 2.85\\
    \multicolumn{2}{l|}{TraPHic \cite{ChandraBhattacharyaEtAl2019}} & - & - & - & 0.64 & - & 2.76\\
    \multicolumn{2}{l|}{Graph-LSTM \cite{ChandraGuanEtAl2020}} & - & - & - & 0.62 & - & 2.45\\
    \hline 
    \multirow{3}{*}{T++ \cite{SalzmannIvanovicEtAl2020}}
    & $\Delta\text{ESV}_1$ & 
    \underline{-0.19} & \underline{-0.15} & \underline{-0.12} \Tstrut \\
    & $\Delta\text{ESV}_2$ & 
    -0.48 & -0.45 & -0.41 & 
    \underline{0.46} & \underline{1.20} & 2.34\\
    & $\Delta\text{ESV}_3$ & 
    -0.53 & -0.50 & -0.46 \\
    \cline{2-8}

    \multirow{3}{*}{$D_{SD}$ Only}
    & $\Delta\text{ESV}_1$ & 
    0.45 & 0.41 & 0.38 \Tstrut \\
    & $\Delta\text{ESV}_2$ & 
    \textbf{0.12} & \textbf{0.11} & \textbf{0.09} & 
    0.69 & 1.36 & \underline{2.28}\\
    & $\Delta\text{ESV}_3$ & 
    \textbf{0.01} & \textbf{8e-3} & \textbf{-2e-3} \\
    \midrule 

    \multirow{3}{*}{Ours}
    & $\Delta\text{ESV}_1$ & 
    \textbf{-0.03} & \textbf{-0.01} & \textbf{7e-3} \\
    & $\Delta\text{ESV}_2$ & 
    \underline{-0.37} & \underline{-0.33} & \underline{-0.29} & 
    \textbf{0.45} & \textbf{1.11} & \textbf{2.12}\\
    & $\Delta\text{ESV}_3$ & 
    \underline{-0.47} & \underline{-0.42} & \underline{-0.37} \\
    \bottomrule
    \end{tabular}
    \label{tab:nuscenes_quant}
    
    \vspace{-0.5cm}
    
\end{table}

\begin{figure*}[t]
    \centering
    \raisebox{-0.5\height}{\begin{overpic}[width=0.28\linewidth,frame]{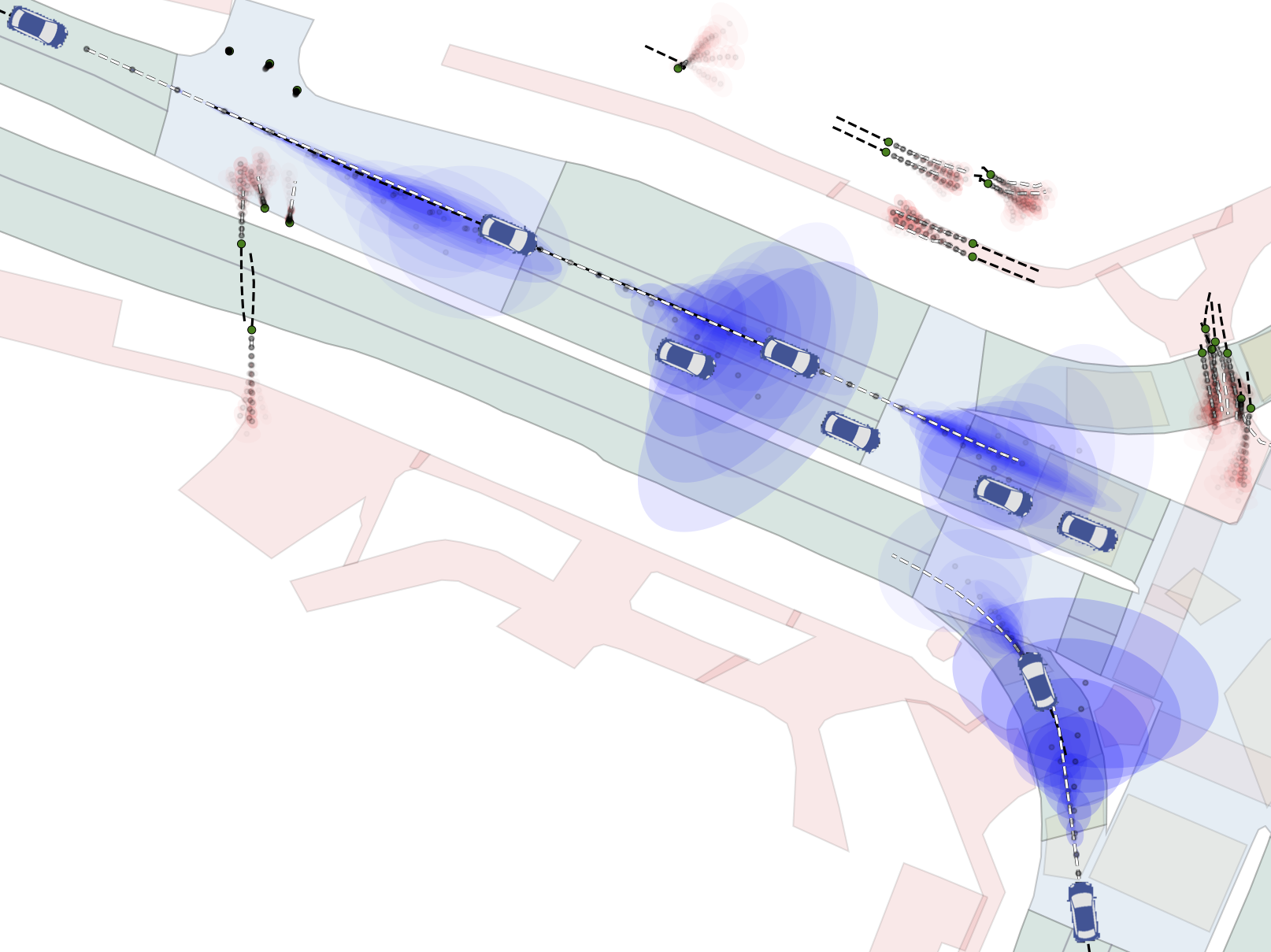}%
    \put (5, 5) {(a) T++}%
    \end{overpic}}
    \raisebox{-0.5\height}{\begin{overpic}[width=0.28\linewidth,frame]{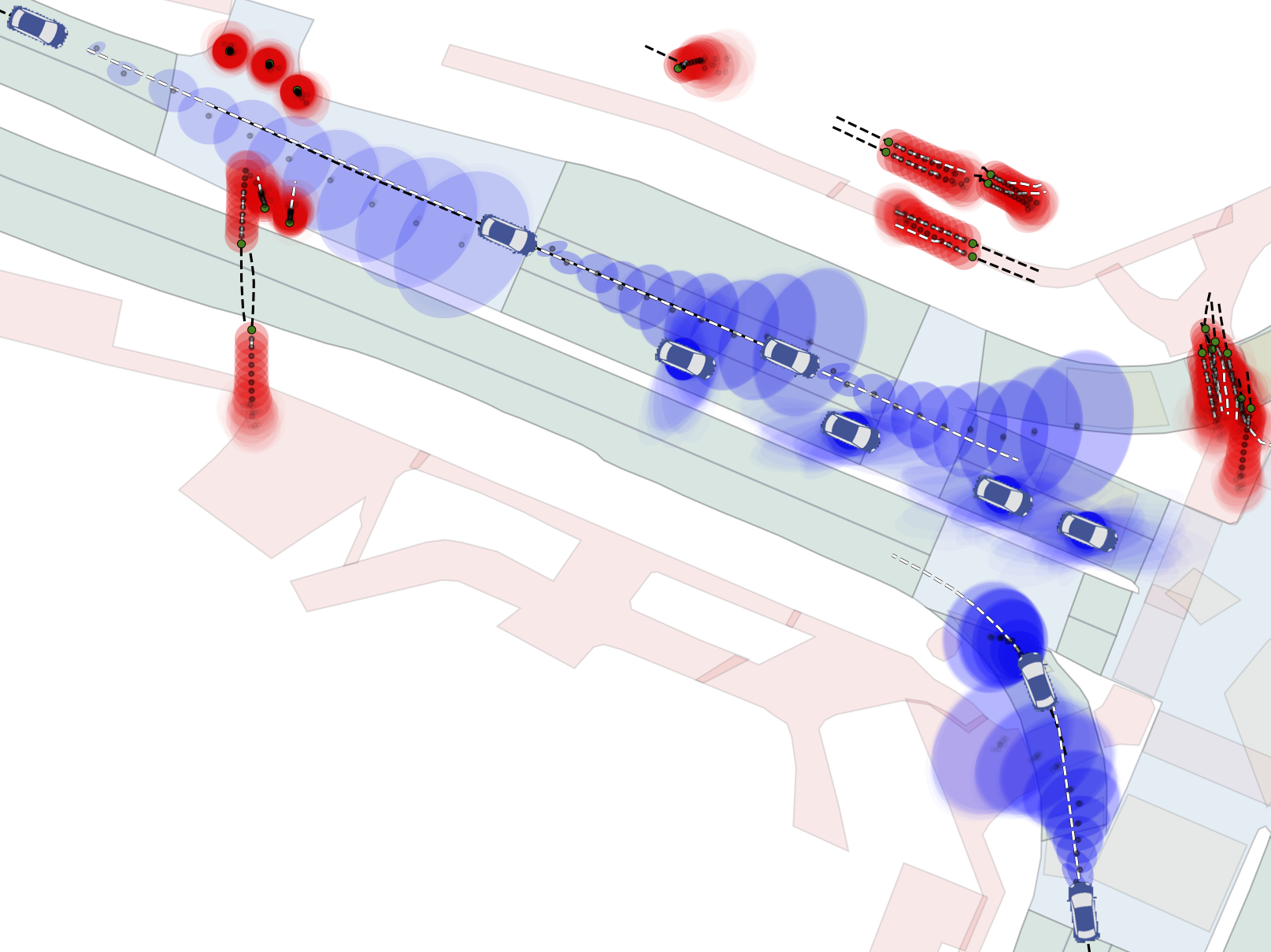}%
    \put (5, 5) {(b) $D_{SD}$ Only}%
    \end{overpic}}
    \raisebox{-0.5\height}{\begin{overpic}[width=0.28\linewidth,,frame]{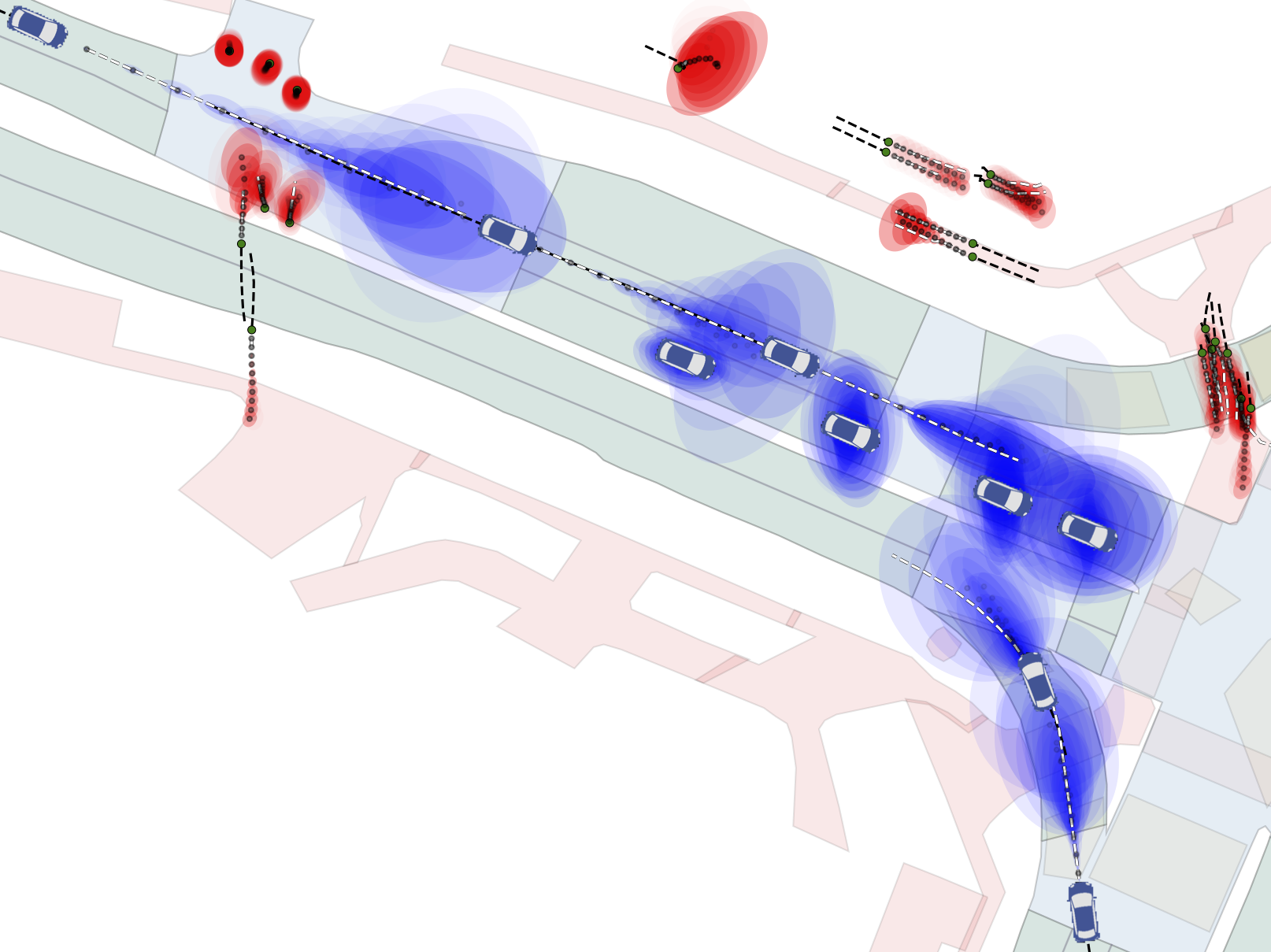}%
    \put (5, 5) {(c) Ours}%
    \end{overpic}}
    \raisebox{-0.5\height}{\begin{overpic}[width=0.1\linewidth,]{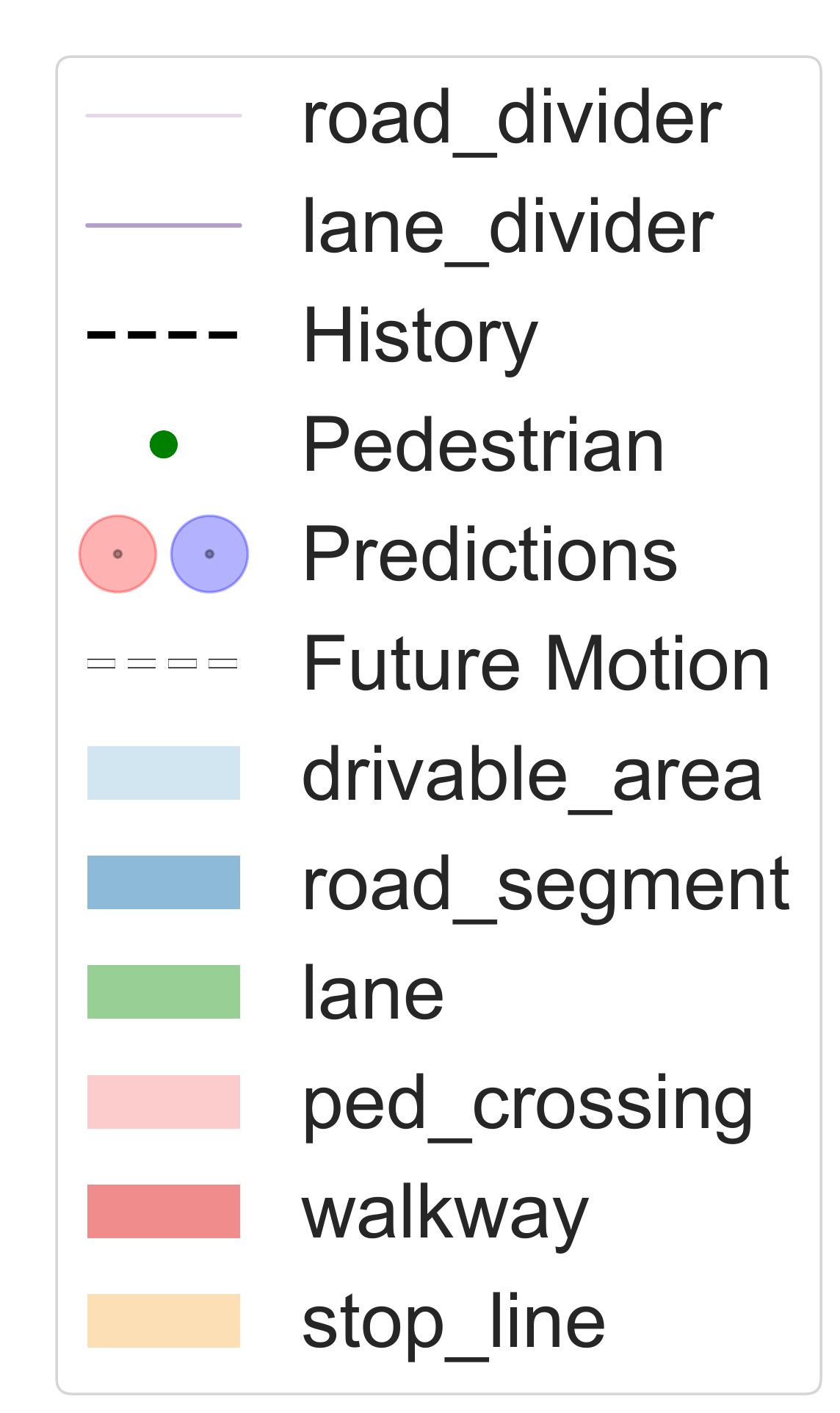}%
    \end{overpic}}
    \caption{Our model's predictions on the nuScenes dataset (with EKF-tracked GT detections) are both more accurate and better calibrated than existing state-of-the-art prediction methods, producing uncertainties that grow sensibly with time and closely align with the GT future.}
    \label{fig:nuscenes_qual}
\end{figure*}

\begin{figure*}[t]
    \centering
    \raisebox{-0.5\height}{\begin{overpic}[width=0.28\linewidth,frame]{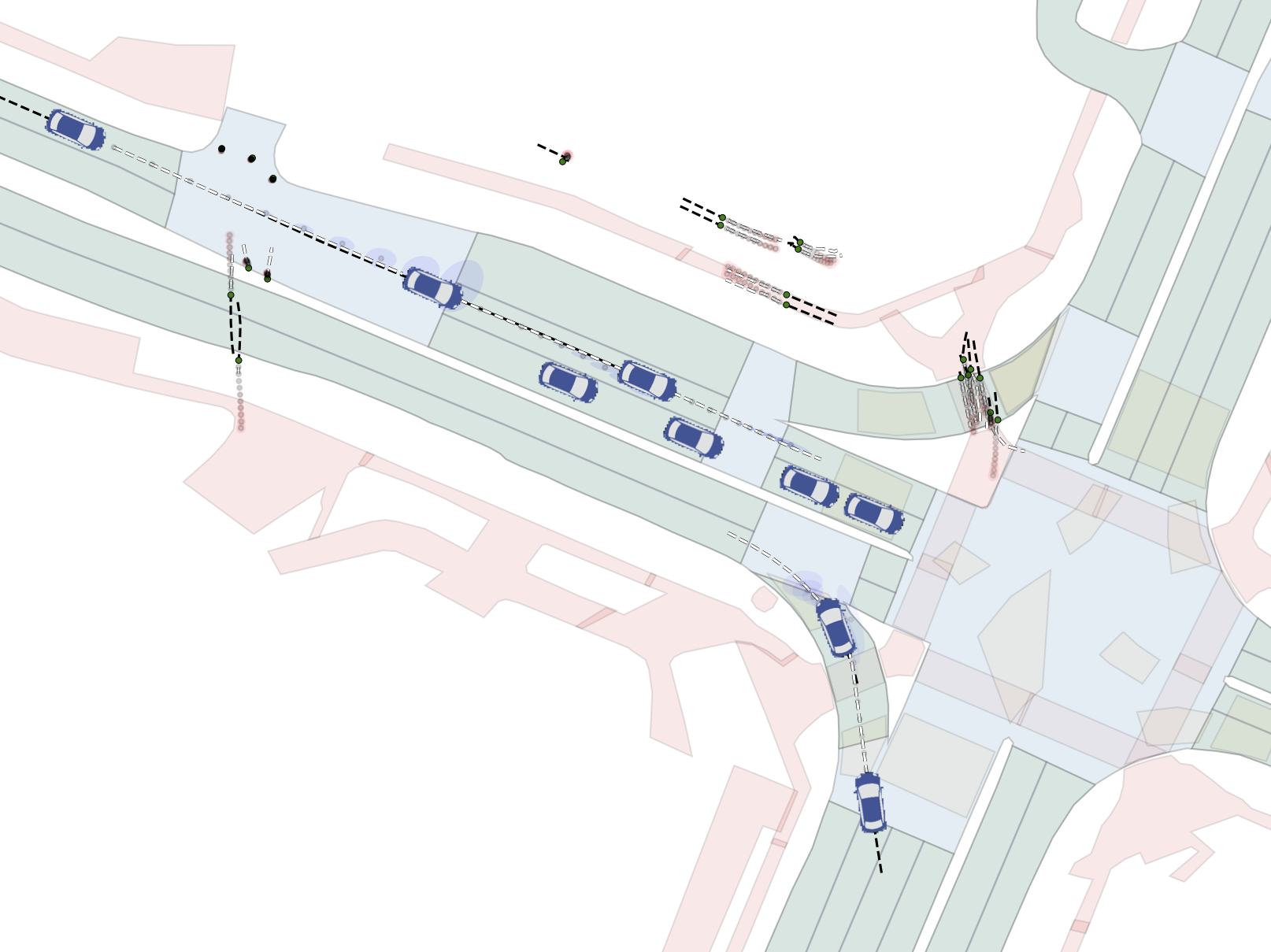}%
    \put (5, 5) {(a) T++}%
    \end{overpic}}
    \raisebox{-0.5\height}{\begin{overpic}[width=0.28\linewidth,frame]{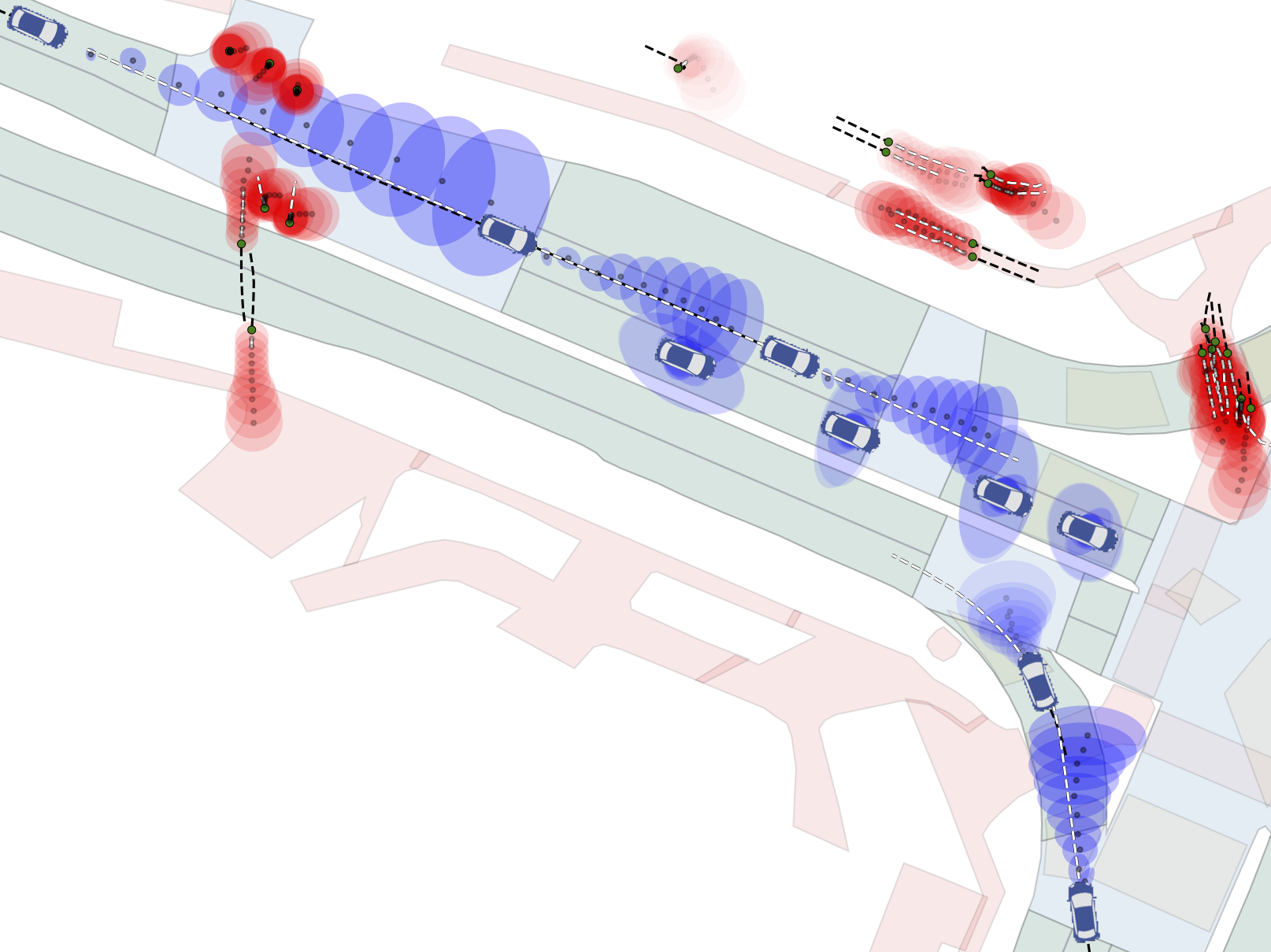}%
    \put (5, 5) {(b) $D_{SD}$ Only}%
    \end{overpic}}
    \raisebox{-0.5\height}{\begin{overpic}[width=0.28\linewidth,,frame]{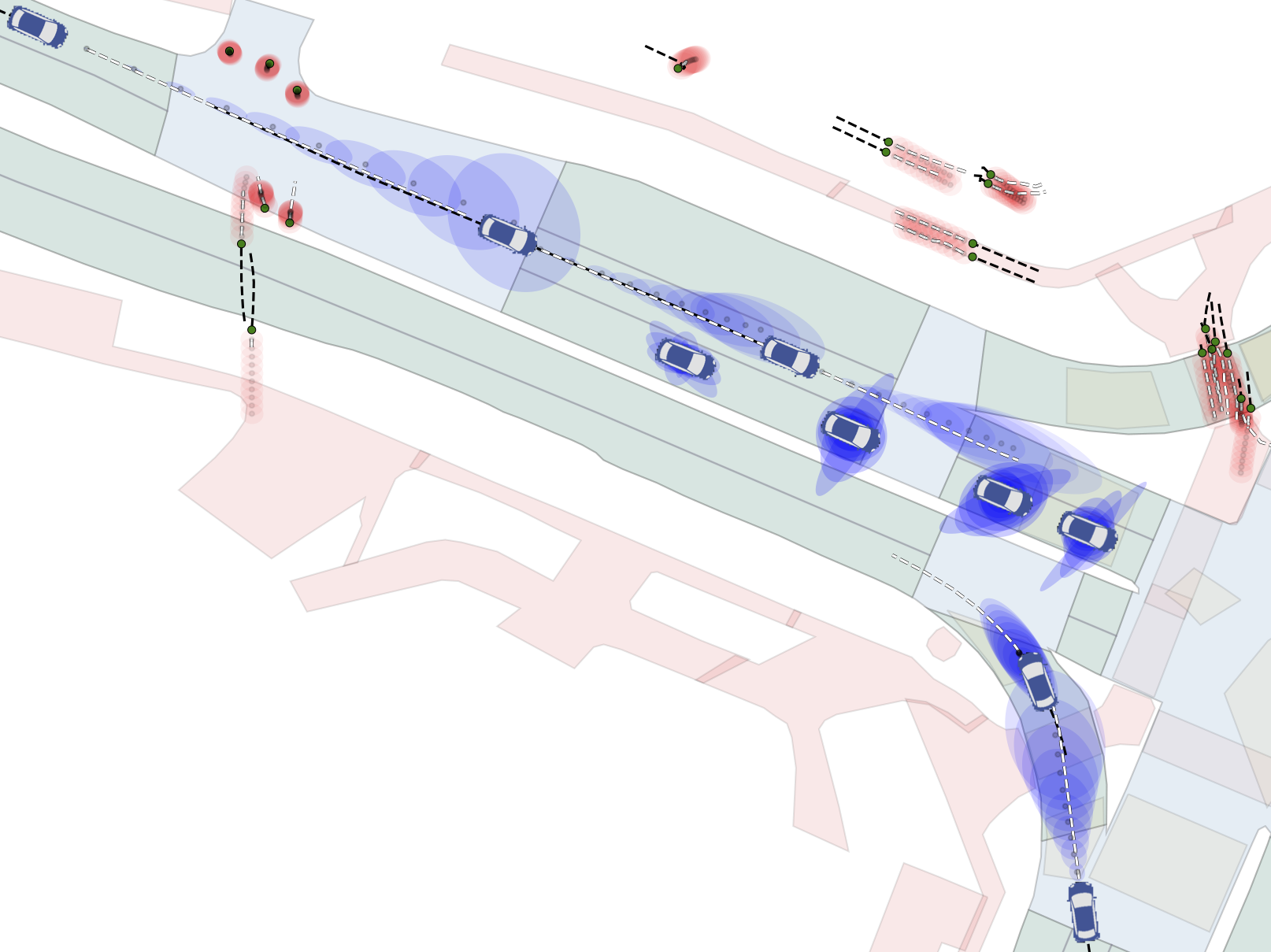}%
    \put (5, 5) {(c) Ours}%
    \end{overpic}}
    \raisebox{-0.5\height}{\begin{overpic}[width=0.1\linewidth,]{figures/nuscenes_legend_v2.pdf}%
    \end{overpic}}
    \caption{We applied the state-of-the-art CenterPoint detector \cite{YinZhouEtAl2021} and AB3DMOT tracker \cite{WengWangEtAl2020} on raw nuScenes LIDAR data to evaluate the performance of our work in a realistic perception scenario.
    Our method's outputs are still the most accurate with sensible uncertainties, due to its propagation of uncertainty.}
    \label{fig:nuscenes_tracked_qual}
    
    \vspace{-0.5cm}
    
\end{figure*}

 \begin{table}[t]
    \centering
    \caption{nuScenes detected with CenterPoint \cite{YinZhouEtAl2021} and tracked by AB3DMOT \cite{WengWangEtAl2020}. Bolded is best, underlined is 2nd-best.}
    \begin{tabular}{l|ccc}
    \toprule
    \multicolumn{1}{c|}{\textbf{Vehicles}} & \multicolumn{3}{c}{\textbf{NLL (nats)}}\\ \cline{1-4}
    \multicolumn{1}{c|}{\textbf{Horizon}} & 1s & 2s & 3s \Tstrut \\
    \midrule 
    T++ \cite{SalzmannIvanovicEtAl2020} & 
    \underline{-0.77}\tiny{$\pm4.23$} & 
    \underline{-0.01}\tiny{$\pm4.49$} & 
    \underline{0.61}\tiny{$\pm4.70$}\\

    $D_{SD}$ Only & 
    1.89\tiny{$\pm0.48$}& 
    2.45\tiny{$\pm1.02$}& 
    3.07\tiny{$\pm1.61$}\\
    \hline

    Ours & 
    \textbf{-1.09}\tiny{$\pm3.09$}&
    \textbf{-0.19}\tiny{$\pm3.38$}&
    \textbf{0.57}\tiny{$\pm3.62$}\Tstrut \\
    \end{tabular}

    \setlength\tabcolsep{4pt}
    \begin{tabular}{l|c|ccc|ccc}
    \toprule
    \multicolumn{2}{c|}{} & \multicolumn{3}{c|}{\textbf{$\Delta$ESV$_i$}} & \multicolumn{3}{c}{\textbf{FDE (m)}} \\ \cline{1-8}
    \multicolumn{2}{c|}{\textbf{Horizon}} & 1s & 2s & 3s & 1s & 2s & 3s \Tstrut \\
    \midrule 
    \multirow{3}{*}{T++ \cite{SalzmannIvanovicEtAl2020}}
    & $\Delta\text{ESV}_1$ & 
    \underline{-0.35} & -0.35 & -0.35 \\
    & $\Delta\text{ESV}_2$ & 
    -0.73 & -0.72 & -0.71 & 
    1.02 & 2.15 & 3.66\\
    & $\Delta\text{ESV}_3$ & 
    -0.81 & -0.79 & -0.78 \\
    \cline{2-8}

    \multirow{3}{*}{$D_{SD}$ Only}
    & $\Delta\text{ESV}_1$ & 
    0.36 & \underline{0.31} & \underline{0.27} \Tstrut \\
    & $\Delta\text{ESV}_2$ & 
    \textbf{0.11} & \textbf{0.06} & \textbf{5e-3} & 
    \underline{1.00} & \underline{1.96} & \underline{3.13}\\
    & $\Delta\text{ESV}_3$ & \
    \textbf{7e-3} & \textbf{-0.01} & \textbf{-0.04} \\
    \midrule 

    \multirow{3}{*}{Ours}
    & $\Delta\text{ESV}_1$ & 
    \textbf{-0.04} & \textbf{-0.03} & \textbf{-0.03} \\
    & $\Delta\text{ESV}_2$ & 
    \underline{-0.35} & \underline{-0.32} & \underline{-0.31} & 
    \textbf{0.78} & \textbf{1.72} & \textbf{3.09}\\
    & $\Delta\text{ESV}_3$ & 
    \underline{-0.38} & \underline{-0.34} & \underline{-0.32} \\
    \bottomrule
    \end{tabular}
    \label{tab:nuscenes_tracked_quant}
    
    \vspace{-0.5cm}
    
\end{table}

\textbf{Qualitative Results.}
In \cref{fig:nuscenes_qual}, we can see that T++ generates almost-invisible uncertainties for its first few predictions which then grow to very large, multi-lane (even multi-road) uncertainties. 
$D_{SD}$ Only limits uncertainty growth, but its predictions veer off into adjacent lanes or walkways. In contrast, our method generates sensible uncertainties that stay within adjacent lanes and accurately cover the GT trajectory.

\textbf{Realistic Perception Stack Results.} To evaluate the performance of our method in a realistic setting (i.e., not using GT inputs), we implemented a perception system based on a state-of-the-art detector and tracker. In particular, we use the CenterPoint detector~\cite{YinZhouEtAl2021} and AB3DMOT tracker~\cite{WengWangEtAl2020} to obtain agent tracks from nuScenes' raw LIDAR data on which our method and its ablations are trained and evaluated. Note that these only affect the inputs to the models, predictions are still compared to GT future agent trajectories. \cref{tab:nuscenes_tracked_quant} summarizes the results, and shows that our method significantly outperforms T++ on FDE while still performing second-best on the $\Delta$ESV$_i$ metrics, mirroring prior trends. \cref{fig:nuscenes_tracked_qual} shows that the same trends from before persist in a realistic perception scenario. Namely, T++ produces overconfident predictions (plotted faintly because T++ predicts multiple low-probability modes and prediction opacity is proportional to mode probability), $D_{SD}$ Only generates underconfident predictions which veer out of lanes or cross sidewalks, whereas our method's predictions stay within lanes and have sensible uncertainty growth that aligns closely with agents' GT future trajectories.

\section{CONCLUSIONS}
In this work, we present a method for incorporating and propagating perceptual state uncertainty through trajectory forecasting. 
At its core, a new statistical-distance based loss function term incentivizes our model to better utilize input uncertainty information and produce more calibrated output trajectory distributions. 
Experiments on both illustrative and real-world datasets show that the added term effectively addresses existing generative trajectory forecasting models' overconfidence, sometimes even improving mean prediction accuracy.

While this work focuses on perceptual state uncertainty, there are many other sources of upstream uncertainty that can be propagated through trajectory forecasting, e.g., agent classification uncertainty and map error, each of which are interesting areas of future work.

\clearpage
\newpage

\addtolength{\textheight}{-17cm}   %

\bibliographystyle{IEEEtran}
\bibliography{ASL_papers,main,added}

\end{document}